\documentclass[11pt,a4paper]{article}
\usepackage{times}
\usepackage{latexsym}

\usepackage{url}
\usepackage{tabularx}

\usepackage{arydshln}
\usepackage{array,etoolbox}
\preto\tabular{\setcounter{magicrownumbers}{0}}
\newcounter{magicrownumbers}

\usepackage{chngpage}
\usepackage{url}
\usepackage{times}  %Required
\usepackage{helvet}  %Required
\usepackage{courier}  %Required
\usepackage{url}  %Required
\usepackage{graphicx}  %Required

\usepackage[utf8]{inputenc}

\usepackage{latexsym}
\usepackage{color}
\usepackage{amsmath}
\usepackage{mathtools}
\usepackage{amssymb}
\usepackage{amsfonts}
\usepackage{breqn}
\usepackage{float}
\usepackage{array}

\usepackage{algorithm}
\usepackage[noend]{algpseudocode}

\usepackage{caption}
\usepackage{subcaption}
\usepackage{paralist}

\usepackage{hyperref}
\usepackage[hyperref]{naaclhlt2019}

\newcommand\enc[1]{\textnormal{enc}\left(#1\right)}
\newcommand\dec[1]{\textnormal{dec}\left(#1\right)}
\newcommand\transl[1]{\texttt{transl}\left(#1\right)}
\newcommand\sli{s_{l_i}}

\newcolumntype{L}[1]{>{\raggedright\let\newline\\\arraybackslash\hspace{0pt}}m{#1}}
\newcolumntype{C}[1]{>{\centering\let\newline\\\arraybackslash\hspace{0pt}}m{#1}}
\newcolumntype{R}[1]{>{\raggedleft\let\newline\\\arraybackslash\hspace{0pt}}m{#1}}

\aclfinalcopy % Uncomment this line for the final submission
\def\aclpaperid{19} %  Enter the acl Paper ID here

\title{Bilingual-GAN: A Step Towards Parallel Text Generation}

\author{Ahmad Rashid, Alan Do-Omri, Md. Akmal Haidar, Qun Liu and Mehdi Rezagholizadeh \\
Huawei Noah's Ark Lab \\
\tt{ahmad.rashid@huawei.com , alan.do.omri@huawei.com}, \\
\tt{md.akmal.haidar@huawei.com, qun.liu@huawei.com}, \\
\tt{mehdi.rezagholizadeh@huawei.com} \\
}

\date{}

\begin{document}
\maketitle
\begin{abstract}
Latent space based GAN methods and attention based sequence to sequence models have achieved impressive results in text generation and unsupervised machine translation respectively. Leveraging the two domains, we propose an adversarial latent space based model capable of generating parallel sentences in two languages concurrently and translating bidirectionally. The bilingual generation goal is achieved by sampling from the latent space that is shared between both languages. First two denoising autoencoders are trained, with shared encoders and  back-translation to enforce a shared latent state between the two languages. The decoder is shared for the two translation directions. Next, a GAN is trained to generate synthetic `code' mimicking the languages’ shared latent space. This code is then fed into the decoder to generate text in either language. We perform our experiments on Europarl and Multi30k datasets, on the English-French language pair, and document our performance using both supervised and unsupervised machine translation.
\end{abstract}

\section{Introduction}

%Machine translation is one of the challenging tasks in Natural Language Processing (NLP). Statistical Machine Translation, proposes phrase-based solutions which requires lots of hand-crafted features. Recently, deep learning opened up a new direction in machine translation which is referred to as neural machine translation (NMT).    
%NMT approaches solve the translation problem end-to-end and they achieve the state-of-the-art performance.  
Many people in the world are fluent in at least two languages, yet most computer applications and services are designed for a monolingual audience. Fully bilingual people do not think about a concept in one language and translate it to the other language but are adept at generating words in either language. 

Inspired by this bilingual paradigm, the success of attention based neural machine translation (NMT) and the potential of Generative Adversarial Networks (GANs) for text generation we propose Bilingual-GAN, an agent capable of deriving a shared latent space between two languages, and then generating from that space in either language. 

Attention based NMT~\citep{bahdanau_attn, convs2s, attn_is_all_you_need} has achieved state of the art results on many different language pairs and is used in production translation systems~\citep{wu2016google}. These systems generally consist of an encoder-decoder based sequence to sequence model where at least the decoder is auto-regressive. Generally, they require massive amount of parallel data but recent methods that use shared autoencoders~\citep{FAE,lample2018phrase} and cross-lingual word embeddings~\citep{muse} have shown promise even without using parallel data.  

Deep learning based text generation systems can be divided into three categories: Maximum Likelihood Estimation (MLE)-based, GAN-based and reinforcement learning (RL)-based. MLE-based methods~\citep{sutskever2014sequence} model the text as an auto-regressive generative process using Recurrent Neural Networks (RNNs) but generally suffer from exposure bias~\citep{scheduled_sampling}. A number of solutions have been proposed including scheduled sampling~\citep{scheduled_sampling}, Gibbs sampling~\citep{gibs_sampling_text_generation} and Professor forcing~\citep{lamb2016professor}.

Recently, researchers have used GANs~\citep{goodfellow2014generative} as a potentially powerful generative model for text~\citep{yu2017seqgan,gulrajani2017improved,Akmal2019b}, inspired by their great success in the field of image generation. Text generation using GANs is challenging due to the discrete nature of text. The discretized text output is not differentiable and if the softmax output is used instead it is trivial for the discriminator to distinguish between that and real text. One of the proposed solutions \citep{ARAE} is to generate the latent space of the autoencoder instead of generating the sentence and has shown impressive results.

We use the concept of shared encoders and multi-lingual embeddings to learn the aligned latent representation of two languages and a GAN that can generate this latent space. Particularly, our contributions are as follows:

\begin{itemize}
    \item We introduce a GAN model, Bilingual-GAN, which can generate parallel sentences in two languages concurrently.
    \item Bilingual-GAN can match the latent distribution of the encoder of an attention based NMT model.
    \item We explore the ability to generate parallel sentences when using only monolingual corpora.
\end{itemize}

\section{Related Work}
\subsection{Latent space based Unsupervised NMT}

A few works \citep{FAE,UNdreaMT,lample2018phrase} have emerged recently to deal with neural machine translation without using parallel corpora, i.e sentences in one language have no matching translation in the other language. The common principles of such systems include learning a language model, encoding sentences from different languages into a shared latent representation and using back-translation \citep{back_translation} to provide a pseudo supervision. \citet{FAE} use a word by word translation dictionary learned in an unsupervised way \citep{facebook_unsup_dict} as part of their back-translation along with an adversarial loss to enforce language independence in latent representations. \citet{lample2018phrase} improves this by removing these two elements and instead use Byte Pair Encoding (BPE) sub-word tokenization \citep{BPE} with joint embeddings learned using FastText \citep{fasttext}, so that the sentences are embedded in a common space. \citet{UNdreaMT} uses online back translation and cross-lingual embeddings to embed sentences in a shared space. They also decouple the decoder so that one is used per language.

\subsection{Latent space based Adversarial Text Generation}
Researchers have conventionally utilized the GAN framework in image applications~\citep{salimans2016improved} with great success. Inspired by their success, a number of works have used GANs in various NLP applications such as machine translation \citep{wu2017adversarial, yang2017improving}, dialogue models~\citep{li2017adversarial}, question answering~\citep{yang2017semi}, and natural language generation~\citep{gulrajani2017improved, kim2017adversarially}. However, applying GAN in NLP is challenging due to the discrete nature of text. Consequently, back-propagation would not be feasible for discrete outputs and it is not straightforward to pass the gradients through the discrete output words of the generator. A latent code based solution for this problem, ARAE, was proposed in ~\citet{kim2017adversarially}, where a latent representation of the text is derived using an autoencoder and the manifold of this representation is learned via adversarial training of a generator. Another version of the ARAE method which proposes updating the encoder based on discriminator loss function was introduced in~\citep{Spinks2018}. \citet{Jules2019} introduced a self-attention based GAN architecture to the ARAE and \citet{Akmal2019a} explore a hybrid approach generating both a latent representation and the text itself.

\section{Methodology}
\begin{figure}[htb!]
 \centering
 \includegraphics[scale=0.23]{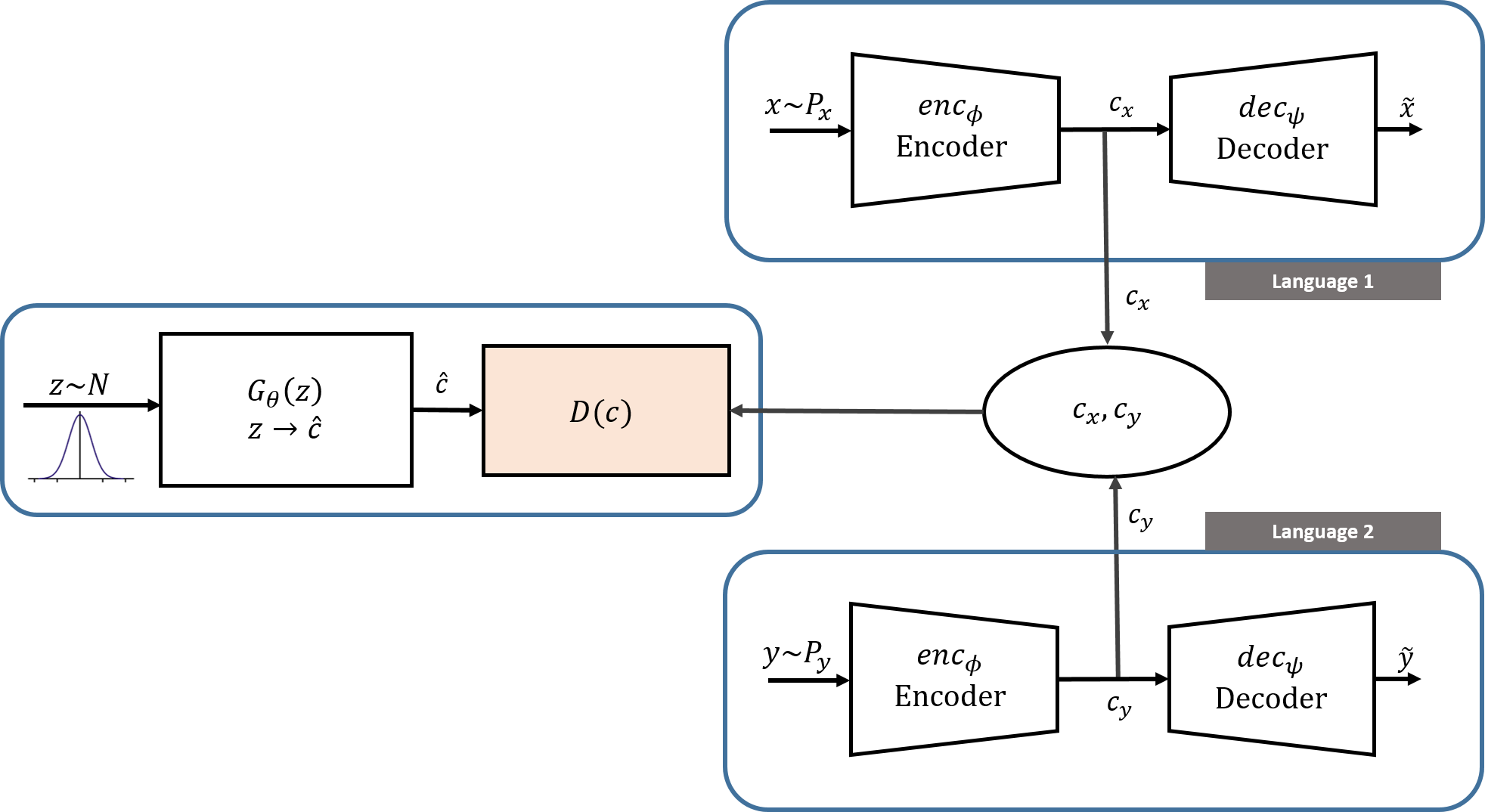}
 \caption{The complete architecture for our unsupervised bilingual text generator (Bilingual-GAN)}\label{fig:BGAN}
 \end{figure}

The Bilingual-GAN comprises of a translation module and a text generation module. The complete architecture is illustrated in Figure~\ref{fig:BGAN}. 
%The middle left rectangle unit represents the text generation unit and the remaining part represents the translation unit.%This section explains both in details including their architecture and their training methods.

\subsection{Translation Unit}
\label{translation}
The translation system is a sequence-to-sequence model with an encoder and a decoder extended to support two languages. This first translation component is inspired by the unsupervised neural machine translation system by \citet{FAE}. We have one corpus in language 1 and another in language 2 (they need not be translations of each other), an encoder and a decoder shared between the two languages. The weights of the encoder are shared across the two languages, only their embedding tables are different. For the decoder, the weights are also shared except for the last language specific projection layer. 

\par The loss function which is used to compare two sentences is the same as the standard sequence-to-sequence loss: the token wise cross-entropy loss between the sentences, that we denote by $\Delta (\textnormal{sentence a}, \textnormal{sentence b})$. For our purpose, let $\sli$ be a sentence in language $i$ with $i \in \{1,2\}$. The encoding of sentence $\sli$ is denoted by $\enc{\sli}$ 
%(the output of the encoder when given the input $\sli$) 
in language $i$ using the word embeddings of language $i$ to convert the input sentence $\sli$. Similarly, denote by $\dec{x, l_i}$ the decoding of the code $x$ (typically the output of the encoder) into language $l_i$ using the word embeddings of target language $i$. 
\par
\begin{figure}[!htb]
\centering
\includegraphics[width=.5\textwidth]{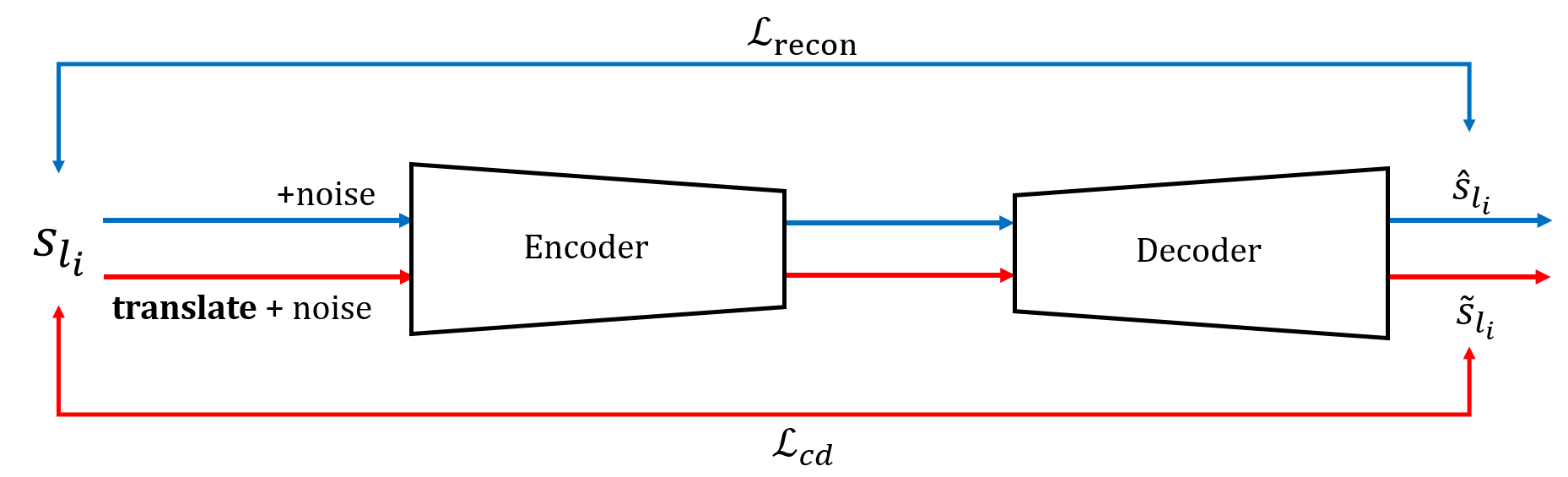}
\caption{The translation unit of the Bilingual-GAN.}\label{fig:fae}
\end{figure}

 Then, the system is trained with three losses aimed to allow the encoder-decoder pair to reconstruct inputs (reconstruction loss), to translate correctly (cross-domain loss) and for the encoder to encode language independent codes (adversarial loss). 
%The losses are applied for every batch for both languages. 
\paragraph{Reconstruction Loss}
This is the standard autoencoder loss which aims to reconstruct the input:
\[ \mathcal{L}_{\textnormal{recon}} = \Delta\left(\sli, \overbrace{\dec{\enc{\sli}, l_i}}^{\mathclap{\hat{s}_{l_i}} \quad \coloneqq}\right) \]
This loss can be seen in Figure \ref{fig:fae}.
\paragraph{Cross-Domain Loss} 
This loss aims to allow translation of inputs. It is similar to back-translation \citep{back_translation}. For this loss, denote by $\transl{\sli}$ the translation of sentence $\sli$ from language $i$ to language $1-i$. The implementation of the translation is explained in subsection \ref{sec:supervision} when we address supervision. 
\begin{equation} 
\mathcal{L}_{\textnormal{cd}} = \Delta\left(\sli, \underbrace{\dec{\enc{\transl{\sli}},l_i}}_{\mathclap{\tilde{s}_{l_i} \; \coloneqq}} \right) \label{eq:cd}
\end{equation}
In this loss, we first translate the original sentence $\sli$ into the other language and then check if we can recreate the original sentence in its original language. This loss can be seen in Figure \ref{fig:fae}.
\paragraph{Adversarial Loss}
\label{sec:adv_loss}
This loss is to enforce the encoder to produce language independent code which is believed to help in decoding into either language. %This loss has been defined adversarially. Let $D$ be a discriminator where $D(c)$ is a prediction for the language of the sentence that was used to create code $c$ (typically the output of an encoder), 0 if the sentence is in language 0 and 1 if the sentence is in language 1. We thus have for the discriminator $D$ the following
%\[ \mathcal{L}_D = \max \{ D(\enc{\sli}) - D(\enc{s_{l_j}}) \} \]
%and for its adversary, the encoder, the opposite:
%\[ \mathcal{L}_{\textnormal{enc}} = \min \{ D(\enc{\sli}) - D(\enc{s_{l_j}}) \} \]
This loss was only present in \citet{FAE} and removed in \citet{lample2018phrase} as it was considered not necessary by the authors and even harmful. Our results show a similar behaviour. 
\paragraph{Input Noise}
In order to prevent the encoder-decoder pair to learn the identity function and to make the pair more robust, noise is added to the input of the encoder. 
%This is illustrated in Figure \ref{fig:fae} where you see the $+$ noise atop the arrows feeding into the encoder. 
On the input sentences, the noise comes in the form of random word drops (we use a probability of 0.1) and of random shuffling but only moving each word by at most 3 positions. We also add a Gaussian noise of mean 0 and standard deviation of 0.3 to the input of the decoder. 
\subsubsection{Supervision}
\label{sec:supervision}
%Recall that in the cross-domain loss above, equation \ref{eq:cd}, the translation function $\transl{\sli}$ was used to translate the sentence $\sli$ from language $i$ to language $1-i$. 
The choice of the translation function $\transl{\sli}$ directly affects the amount of supervision in the trained model. 
%Indeed, notice that only $\sli$ and $\transl{\sli}$ are used in the losses.
If the translation function $\transl{}$ is a lookup of a word-by-word translation dictionary learned in an unsupervised fashion as in \citet{facebook_unsup_dict}, then the whole system is trained in an unsupervised manner since we have no groundtruth information about $\sli$. After a couple of epochs, the encoder-decoder model should be good enough to move beyond simple word-by-word translation. At that point the translation function can be changed to using the model itself to translate input sentences. This is what's done in \citet{FAE} where they change the translation function from word-by-word to model prediction after 1 epoch. In our case, we get the word-by-word translation lookup table by taking each word in the vocabulary and looking up the closest word in the other language in the multilingual embedding space created by \citet{muse}. 
\par If the translation function $\transl{}$ is able to get the ground truth translation of the sentence, for example if we have an aligned dataset, then $\transl{\sli}=s_{l_j}$ which is encoded and decoded into the original language $i$ and compared with $\sli$ getting the usual supervised neural machine translation loss. 
% However, note that this supervision is only one way since you learn to predict in language $i$ given a sentence in language $j$. We refer to this level of supervision as Half-Supervised in our results section later. In order to have supervision both ways, one would need to have both $\sli$ and $s_{l_j}$ in the training corpus, this is what we refer to as the Supervised level. 
\subsubsection{Embeddings}
There are a few choices for embedding the sentence words before feeding into the encoder. 
%We experiment with a few and show the results in section \ref{seq:transl_results}.
In particular, we use randomly initialized embeddings, embeddings trained with FastText \citep{fasttext} and both pretrained and self-trained cross-lingual embeddings \citep{muse}.

\subsection{Bilingual Text Generation Unit }
% The Bilingual-GAN is mainly composed of two parts: an unsupervised machine translation network and an adversarial language generator as depicted in Figure~\ref{fig:BGAN}. For training, a UNMT system \cite{FAE} is first trained and kept frozen thereafter.
% This UNMT system has a single encoder and decoder for the two translation directions, instead of two as would normally be the case in supervised NMT. 
% The input and output languages are dictated by start tokens associated to the selected language in the encoder and the decoder respectively.

%First, we pre-train our NMT system~\ref{translation}. The NMT system learns a shared latent space ($c_x, c_y$) for the two language directions, and this shared latent space is enforced by a GAN setup between a discriminator and the encoders, and through back-translation\citep{back_translation}. Then, a bilingual generator is trained adversarially to learn the manifold of the shared latent space ($c_x, c_y$), which is learned in the NMT system. It is trained similar to a modified version of ARAE~\citep{Spinks2018} to generate codes $\hat{c}$ which mimic the samples from the shared latent space. Once GAN training is finished, the decoders of the NMT system can be used to generate parallel bilingual sentences by decoding the generator output code, $\hat{c}$.
%The complete training algorithm is described in Algorithm~\ref{alg:BL-GAN}. 

%\subsubsection{Neural Text Generation}
The proposed bilingual generator is a GAN trained to learn the latent state manifold of the encoder of the translation unit. We use the Improved Wasserstein GAN gradient penalty (IWGAN)~\citep{gulrajani2017improved} loss function in our experiments:     

\begin{dmath}
L = \mathbb{E}_{\hat{c}\sim\mathbb{P}_{g}}[{D(\hat{c})}] - \mathbb{E}_{c{\sim}\mathbb{P}_{r}}[D(c)] + \lambda\mathbb{E}_{\bar{c}\sim\mathbb{P}_{\bar{g}}}[(||\nabla_{\bar{c}}    D(\bar{c})||_2-1)^2])
%[({\|\nabla{_{\hat{c}}}{D(\hat{c})}\|}_2-1)^2]
\label{wgan_gp}
\end{dmath}
where $\mathbb{P}_r$ is the is the real distribution, $c$ represents the `code' or the latent space representation of the input text, $\mathbb{P}_g$ is the fake or mimicked distribution, $\hat{c}$ represents the generated code representation. The last term is the gradient penalty where $[\bar{c}\sim \mathbb{P}_{\bar{g}}(\bar{c})] \leftarrow \alpha ~ [c\sim \mathbb{P}_{r}(c)] + (1-\alpha)~ [\hat{c}\sim \mathbb{P}_{g}(\hat{c})]$ and it is a  random latent code obtained by sampling uniformly along a line connecting pairs of the generated code and the encoder output.  $\lambda$ is a constant. We used $\lambda = 10$ in our experiments. 

\subsubsection{Matrix-based code representation}
In latent-space based text generation, where the LSTM based encoder-decoder architectures do not use attention, a single code vector is generally employed which summarizes the entire hidden sequence~\citep{ARAE}. A variant of the approach is to employ global mean pooling to produce a representative encoding~\citep{cifka}. We take advantage of our attention based architecture and our bidirectional encoder to concatenate the forward and backward latent states depth-wise and produce a code matrix which can be attended to by our decoder. The code matrix is obtained by concatenating the latent code of each time steps. Consequently, the generator tries to mimic the entire concatenated latent space. We found that this richer representation improves the quality of our sentence generation.

\subsubsection{Training}
First we pre-train our NMT system (see section~\ref{translation}). In order to train the GAN, we used the encoder output of our NMT system as 'real' code. The encoder output is a latent state space matrix which captures all the hidden states of the LSTM encoder. Next we generate noise which is upsampled and reshaped to match the dimensions of the encoder output. This is then fed into a generator neural network comprising 1 linear layer and 5 1-d convolutional with residual connections. Finally we pass it through a non-linearity and output the fake code. The `real' code and the fake code are then fed into the discriminator neural  network, which also consists of $5$ convolutional and $1$ linear layer. The last layer of the discriminator is a linear layer which ouputs a score value. The discriminator output is used to calculate the generator and discriminator losses. The losses are optimized using Adam~\citep{Adam}. Unlike the GAN update in~\cite{gulrajani2017improved}, we use $1$ discriminator update per generator update. We think that because we train our GAN on the latent distribution of machine translation we get a better signal to train our GAN on and don't require multiple discriminator updates to one generator update like in~\citet{ARAE}

In one training iteration, we feed both an English and a French sentence to the encoder and produce two real codes. We generate one fake code by using the generator and calculate losses against both the real codes. We average out the two losses. Although, the NMT is trained to align the latent spaces and we can use just one language to train the GAN, we use both real codes to reduce any biases in our NMT system. We train our GAN on both the supervised and unsupervised NMT scenarios. In the supervised scenario, we feed  English and French parallel sentences in each training iteration. In the unsupervised scenario, our corpus does not contain parallel sentences.

Once the GAN is trained, the generator code can be decoded in either language using the pre-trained decoder of the NMT system.

%Algorithm~\ref{alg:BL-GAN} describes the training for our GAN.  

%We use the Wassertein GAN \cite{arjovsky2017wasserstein} value function to train our model, which in the code space is as follows:
%\[
%\min_{G}\max_{D}\mathbb{E}_{c\sim\mathbb{P}_{r}}[D(c)] - \mathbb{E}_{\hat{c}{\sim}\mathbb{P}_{g}}[D(\hat{c})]
%\]

%where $D:c\mapsto\mathbb{R}$ denotes the discriminator and $\mathbb{P}_{r}$ and $\mathbb{P}_{g}$ are real and generated distributions respectively. 
%If the discriminator parameters are restricted to an 1-Lipshitz function set, this term corresponds to minimizing the Wasserstein 1-Distance. 

%to enforce this by constraining the gradient norm of the discriminator output with respect to its input. 

\section{Experiments}
This section presents the different experiments we did, on both translation and bilingual text generation, and the datasets we worked on. 

% please add the references for the datasets
\subsection{Datasets}
The Europarl and the Multi30k datasets have been used for our experimentation. The Europarl dataset is part of the WMT 2014 parallel corpora~\citep{koehn2005europarl} and contains a little more than 2 millions French-English aligned sentences. The Multi30k dataset is used for image captioning~\citep{multi30k} and consists of 29k images and their captions. We only use the French and English paired captions. 
\par As preprocessing steps on the Europarl dataset, we removed sentences longer than 20 words and those with a ratio of number of words between translations is bigger than 1.5. Then, we tokenize the sentence using the Moses tokenizer \citep{moses}. For the Multi30k dataset, we use the supplied tokenized version of the dataset with no further processing. For the BPE experiments, we use the sentencepiece subword tokenizer by Google \footnote{https://github.com/google/sentencepiece}. Consequentially, the decoder also predicts subword tokens. This results in a common embeddings table for both languages since English and French share the same subwords. The BPE was trained on the training corpora that we created. 
\par For the training, validation and test splits, we used 200k, after filtering, randomly chosen sentences from the Europarl dataset for training and 40k sentences for testing. When creating the splits for unsupervised training, we make sure that the sentences taken in one language have no translations in the other language's training set by randomly choosing different sentences for each of them with no overlap. For the validation set in that case, we chose 80k sentences. In the supervised case, we randomly choose the same sentences in both languages with a validation set of 40k. For the Multi30k dataset, we use 12\,850 and 449 sentences for training and validation respectively for each language for the unsupervised case and the whole provided split of 29k and 1014 sentences for training and validation respectively in the supervised case. In both cases, the test set is the provided 1k sentences Flickr 2017 one. 
For the hyperparameter search phase, we chose a vocabulary size of 8k for the Europarl, the most common words appearing in the training corpora and for the final experiments with the best hyperparameters, we worked with a vocabulary size of 15k. For Multi30k, we used the 6800 most common words as vocabulary. 

\subsection{System Specifications}
\textbf{NMT Unit} The embeddings have size 300, the encoder consists of either 1 or 2 layers of 256 bidirectional LSTM cells, the decoder is equipped with attention \citep{bahdanau_attn} and consists of a single layer of 256 LSTM cells. The discriminator, when the adversarial loss is present, is a standard feed-forward neural network with 3 layers of 1024 cells with ReLU activation and one output layer of one cell with Sigmoid activation. 
\par We used Adam with a $\beta_1$ of 0.5, a $\beta_2$ of 0.999, and a learning rate of 0.0003 to train the encoder and the decoder whereas we used RMSProp with a learning rate of 0.0005 to train the discriminator. Most of the specifications here were taken from \citet{FAE}.

\textbf{NTG Unit} The Generator and Discriminator are trained using Adam with a a $\beta_1$ of 0.5, a $\beta_2$ of 0.999, and a learning rate of 0.0001. 

\subsection{Quantitative Evaluation Metrics}
%\subsubsection{BLEU}
%\subsubsection{Generation BLEU}

\textbf{Corpus-level BLEU} We use the BLEU-N scores to evaluate the fluency of the generated sentences according to~\citet{papineni2002},   
\begin{equation}
\text{BLEU-N} = \text{BP} \cdot \text{exp}(\sum_{n=1}^N w_n \text{log} (p_n) )\\
\end{equation}  
where $p_n$ is the probability of $n$-gram and $w_n = \frac{1}{n}$.
The results is described in Table~\ref{bleu-n}.
Here, we set \textit{BP} to 1 as there is no reference length like in machine translation. For the evaluations, we generated 40\,000 sentences for the model trained on Europarl and 1\,000 on the model trained on Multi30k.

\textbf{Perplexity} is also used to evaluate the fluency of the generated sentences. For the perplexity evaluations, we generated 100\,000 and 10\,000 sentences for the Europarl and the Multi30k datasets respectively. The forward and reverse perplexities of the LMs trained with maximum sentence length of 20 and 15 using the Europarl and he  Multi30k datasets respectively are described in Table~\ref{ppl}. The forward perplexities (F-PPL) are calculated by training an RNN language model (RNNLM)~\citep{zaremba2015} on real  training  data  and  evaluated  on  the generated samples. This measure describe the fluency of the synthetic samples. We also calculated  the  reverse  perplexities (R-PPL) by  training  an RNNLM on the synthetic samples and evaluated on the real test data. The results are illustrated in Table~\ref{ppl}.
%We can easily compared the performance of the LMs by using the forward perplexities while it is not possible  by  using  the  reverse  perplexities  as  the  models  are trained using the synthetic samples with different vocabulary sizes

\subsection{Translation} \label{seq:transl_results} 
%We used many hyper-parameters in our experiments and to keep the results table compact, we abbreviated a few.
\begin{table}[!htb]
    \centering
  \scalebox{0.85}{
  \begin{tabular}{|l|l|}
    \hline
     \textbf{MTF}  & the epoch at which we stop using the  \\ 
     & $\transl{}$ function and instead start using  \\
     & the model \\
     \hline
     \textbf{NC} & a new concatenation method used to combine \\ 
     & the bidirectional encoder output: \\ 
     & concatenate either the forward and backward \\
     & states lengthwise or depthwise\\
     \hline
     \textbf{FastText}& the use of FastText \citep{fasttext} \\
     &to train our embeddings\\
     \hline
     \textbf{Xlingual} & refers to the use of cross-lingual embeddings \\
     & using \cite{muse} either trained  \\
     & on our own (\textbf{Self-Trained}) or pretrained \\
     &  (\textbf{Pretrain.}) ones.\\
     \hline
     \textbf{BPE}& the use of subword tokenization learned as \\
     & in \cite{BPE}\\
     \hline
     \textbf{NoAdv}& not using the adversarial loss to train the \\
     & translation part described in section \ref{sec:adv_loss}\\
     \hline
     \textbf{2Enc}& using a 2 layers of 256 cells each bidirectional\\ 
     &LSTM encoder\\
     \hline

  \end{tabular}}
  \caption{Notations that are used for this experiment}
  \label{notation}
\end{table}
This section of the results focuses on the scores we have obtained while training the neural machine translation system. The results in Table~\ref{tab1} will show the BLEU scores for translation on a held out test set for the WMT'14 Europarl corpus and for the official Flickr test set 2017 for the Multi30k dataset. The notations that are used in Table~\ref{tab1} are described in Table~\ref{notation}.
\begin{table*}[!htb]
\begin{adjustwidth}{-1in}{-1in}% adjust the L and R margins by 1 inch
\centering
\footnotesize
% \begin{tabular}{@{\makebox[1.5em][r]{}}|L{11cm}|c|c|c|} 
\begin{tabular}{|L{11cm}|c|c|c|} 
\multicolumn{4}{c}{\textbf{Europarl}} \\ \hline
& \textbf{FR to EN} & \textbf{EN to FR} & \textbf{Mean} \\ \hline
\textbf{Supervised + Train. Pretrain. Xlingual + NC + 2Enc + NoAdv*} &  \textbf{26.78} & \textbf{26.07} & \textbf{26.43} \\ \hline
% Supervised + Train. Pretrain. Xlingual + NC + MTF 5 + 2Enc + NoAdv & 24.79 & 25.08 & 24.94 \\ \hline
Supervised + NC & 24.43 & 24.89 & 24.66 \\ \hline
% \textbf{Half-Supervised + NoAdv + NC} & \textbf{27.79} & \textbf{26.59} & \textbf{27.19} \\ \hline
% Half-Supervised + 2Enc & 26.49 & 26.00 & 26.25 \\ \hline
% Half-Supervised + NC & 24.56 & 24.44 & 24.50 \\ \hline
% \textbf{Half-Supervised Vanilla} & \textbf{23.15} & \textbf{23.76} & \textbf{23.46} \\ \hline
% Half-Supervised + BPE & 23.96 & 13.00 & 18.48 \\ \hline
% Unsupervised + Train. Self-Trained FastText Embeddings + NC + MTF 5 & 18.12 & 17.74 & 17.93 \\ \hline
% Unsupervised + Train. Pretrain. Xlingual + NC + MTF 5 & 17.42 & 17.34 & 17.38 \\ \hline
\textbf{Unsupervised + Train. Pretrain. Xlingual + NC + MTF 5 + 2Enc + NoAdv*} & \textbf{20.82} & \textbf{21.20} & \textbf{21.01} \\ \hline
Unsupervised + Train. Self-Trained FastText Embeddings + NC + MTF 5 & 18.12 & 17.74 & 17.93 \\ \hline
Unsupervised + Train. Pretrain. Xlingual + NC + MTF 5 & 17.42 & 17.34 & 17.38 \\ \hline
Unsupervised + NC + MTF 4 & 16.45 & 16.56 & 16.51 \\ \hline
Unsupervised + Train. Self-Trained Xlingual + NC + MTF 5 & 15.91 & 16.00 & 15.96 \\ \hline
\textbf{Baseline} (Unsupervised + Fixed Pretrain. Xlingual + NC + MTF 5) & 15.22 & 14.34 & 14.78 \\ \hline
\multicolumn{4}{c}{\textbf{Multi30k}} \\ \hline
\textbf{Supervised + Train. Pretrain. Xlingual + NC + 2Enc + NoAdv} & \textbf{36.67} & \textbf{42.52} & \textbf{39.59} \\ \hline
% Supervised + Train. Pretrain. Xlingual + NC + MTF 5 + 2Enc + NoAdv & 30.12 & 35.17 & 32.65 \\ \hline
\textbf{Unsupervised + Train. Pretrain. Xlingual + NC + MTF 5 + 2Enc + NoAdv} & \textbf{10.26} & \textbf{10.98} & \textbf{10.62} \\ \hline

\end{tabular}
\end{adjustwidth}
\caption{The BLEU-4 scores for French to English and English to French translation. The *'ed experiments use a vocabulary size of 15k words. The Multi30k experiments use the best hyperparameters found when training on the Europarl dataset and a vocabulary size of 6800 words.}
\label{tab1}
\end{table*}
The baseline is our implementation of the architecture from \citet{FAE}.  From Table~\ref{tab1}, we notice first that removing the adversarial loss helps the model. It's possible that the shared encoder and decoder weights are enough to enforce a language independent code space. We note that using 2 layers for the encoder is beneficial but that was to be expected. We also note that the  new concatenation method improved upon the model. A small change for a small improvement that may be explained by the fact that both the forward and the backward states are combined and explicitly represent each word of the input sentence rather than having first only the forward states and then only the backward states. 
\par Surprisingly, BPE gave a bad score on English to French. We think that this is due to French being a harder language than English but the score difference is too big to explain that. Further investigation is needed. We see also good results with trainable FastText embeddings trained on our training corpora. Perhaps using pre-trained ones might be better in a similar fashion as pre-trained cross-lingual embeddings helped over the self-trained ones. The results also show the importance of letting the embeddings change during training instead of fixing them. 

\subsection{Text Generation}
We evaluated text generation on both the fluency of the sentences in English and French and also on the degree to which concurrently generated sentences are valid translations of each other. We fixed our generated sentence length to a maximum of length 20 while training on Europarl and to a maximum of length 15 while training on Multi30k. We measured our performance both on the supervised and unsupervised scenario. The supervised scenario uses a pre-trained NMT trained on parallel sentences and unsupervised uses a pre-trained NMT trained on monolingual corpora. The baseline is our implementation of \citet{ARAE} with two additions. We change the Linear layers to 1-d convolutions with residual connections and our generator produces a distributed latent representation which can be paired with an attention based decoder.

Corpus-level BLEU scores are measured using the two test sets.  %These numbers match the size of the test sets. 
The results are described in Table~\ref{bleu-n}.
\begin{table*}[!htb]
  
  \small
  
  \centering
  \begin{tabular}{|l|l|l|l|l|l|}
  \multicolumn{5}{c}{\textbf{Europarl}} \\ \hline
    %\toprule
    %\multicolumn{3}{c}{Part}                   \\
    %\cmidrule{1-2}
     &  \multicolumn{3}{c|}{\textbf{English}}&\multicolumn{2}{|c|}{\textbf{French}} \\ \hline
           &Bilingual-GAN &Bilingual-GAN & Baseline&Bilingual-GAN &Bilingual-GAN    \\
           &(Supervised) &(Unsupervised) & (ARAE)& (Supervised) &(Unsupervised)   \\
           
    \hline

    %\midrule
    \textit{B-2}           &89.34      &86.06  & 88.55  &82.86    &77.40\\
    \textit{B-3}           &73.37      &70.52  & 70.79 &65.03    &58.32\\
    \textit{B-4}           &52.94      &50.22  & 48.41  &44.87    &38.70\\
    \textit{B-5}           &34.26      &31.63  &  29.07 &28.10    &23.63\\
    \hline
    \multicolumn{5}{c}{\textbf{Multi30k}} \\ \hline
    \textit{B-2}  &68.41 &68.36 & 72.17 &60.23 &61.94\\
    \textit{B-3}  &47.60 &47.69 & 51.56 &41.31 &41.76\\
    \textit{B-4}  &29.89 &30.38 & 33.04 &25.24 &25.60\\
    \textit{B-5}  &17.38 &18.18 & 19.31 &14.21 &14.52\\
    \hline
    %\bottomrule
  \end{tabular}
  \caption{Corpus-level BLEU scores for Text Generation on Europarl and Multi30k Datasets}
  \label{bleu-n}
\end{table*}
The higher BLEU scores demonstrate that the GAN can  generate fluent sentences both in English and French. We can note that the English sentences have a higher BLEU score which could be a bias from our translation system. On Europarl our BLEU score is much higher than the baseline indicating that we can improve text generation if we learn from the latent space of translation rather than just an autoencoder. This however, requires further investigation. 
The BLEU scores for the Multi30k are lower because of the smaller test size. 

Perplexity result is presented in Table~\ref{ppl}. 
%The forward perplexity results are obtained by training a recurrent neural network language model (RNNLM)~\cite{zaremba2015} using the real data and evaluated on the synthetic sentences. They  describe the fluency of the synthetic sentences. We also report the reverse perplexities which are calculated by training RNNLMs using the synthetic sentences and evaluate them on real test data.
We can easily compare different models by using the forward perplexities whereas it is not possible by using the reverse perplexities as the models are trained using the synthetic sentences with different vocabulary sizes. We put the baseline results only for the English generated sentences to show the superiority of our proposed Bilingual generated sentences. The forward perplexities (F-PPL) of the LMs using real data are 140.22 (En), 136.09 (Fr) and 59.29 (En), 37.56 (Fr) for the Europarl and the Multi30k datasets respectively reported in F-PPL column. From the tables, we can note the models with lower forward perplexities (higher fluency) for the synthetic samples tend to have higher reverse perplexities.  For the Europarl dataset, the lower forward perplexities for the Bilingual-GAN and the baseline models than the real data indicate the generated sentences by using these models has less diversity than the training set . For the Multi30k dataset, we cannot see this trend as the size of the test set is smaller than the number of synthetic sentences.
\begin{table*}[!htb]
    \small
    \centering
  \begin{tabular}{|l|l|l|l|l|}
  \multicolumn{5}{c}{\textbf{Europarl}} \\ \hline
    %\toprule
    %\multicolumn{3}{c}{Part}                   \\
    %\cmidrule{1-2}
     &  \multicolumn{2}{c|}{\textbf{English}}&\multicolumn{2}{|c|}{\textbf{French}} \\ \hline
           &\textit{F-PPL} &\textit{R-PPL} &\textit{F-PPL} &\textit{R-PPL}   \\
    \hline

    %\midrule
       Real   & 140.22& - & 136.09& -\\
      Bilingual-GAN (Supervised) &  64.91 & 319.32 & 66.40 &  428.52    \\
      %Real (Unsupervised) &144.16 & - & 140.25  & - \\
      Bilingual-GAN (Unsupervised) & 65.36 & 305.96 & 82.75 & 372.27 \\
      Baseline (ARAE) & 73.57 & 260.18 & - & - \\
      \hline
      
    \multicolumn{5}{c}{\textbf{Multi30k}} \\ \hline
      Real     & 59.29& - & 37.56& -\\
      Bilingual-GAN (Supervised) &  65.97 & 169.19 & 108.91 & 179.12\\
      %Real (Unsupervised) &58.52 & - & 43.45  & - \\
      Bilingual-GAN (Unsupervised) & 83.49 & 226.16 & 105.94 & 186.97 \\
      Baseline (ARAE) & 64.4 & 222.89 & - & - \\
      \hline

    %\bottomrule
  \end{tabular}
  \caption{Forward (F) and Reverse (R) perplexity (PPL) results for the Europarl and Multi30k datasets using synthetic sentences of maximum length 20 and 15 respectively. F-PPL: Perplexity of a language model trained on real data and evaluated on synthetic samples. R-PPL: Perplexity of a language model trained on the synthetic samples from Bilingual-GAN and evaluated on the real test data.
  }
  \label{ppl}
\end{table*}

\begin{table*}[!htb]
\begin{adjustwidth}{-1in}{-1in}
\centering
\small
\begin{tabular}{C{7cm}|C{7cm}} 
\textbf{English} & \textbf{French} \\ \hline
\multicolumn{2}{c}{\textbf{Europarl Supervised}} \\ \hline
the vote will take place tomorrow at 12 noon tomorrow. & le vote aura lieu demain à 12 heures. \\ \hline
mr president, i should like to thank mr. unk for the report. & monsieur le président, je tiens à remercier tout particulièrement le rapporteur.\\ \hline
i think it is now as a matter of trying to make it with a great political action. & je pense dès lors qu'une deuxième fois, je pense que nous pouvons agir à une bonne manière que nous sommes une bonne politique. \\ \hline
the debate is closed. & le débat est clos. \\ \hline
\multicolumn{2}{c}{\textbf{Europarl Unsupervised}} \\ \hline
the report maintains its opinion, the objective of the european union. & la commission maintient son rapport de l ’ appui, tout son objectif essentiel. \\ \hline
the question is not on the basis of which the environmental application which we will do with. & le principe n'est pas sur la loi sur laquelle nous avons besoin de l'application de la législation. \\ \hline
i have no need to know that it has been adopted in a democratic dialogue. & je n'ai pas besoin de ce qu'il a été fait en justice. \\
\hline
\multicolumn{2}{c}{\textbf{Multi30k Supervised}} \\ \hline

a child in a floral pattern, mirrored necklaces, walking with trees in the background. & un enfant avec un mannequin, des lunettes de soleil, des cartons, avec des feuilles. \\ \hline
two people are sitting on a bench with the other people. & deux personnes sont assises sur un banc et de la mer. \\ \hline
a man is leaning on a rock wall. & un homme utilise un mur de pierre. \\ \hline
a woman dressed in the rain uniforms are running through a wooden area & 
une femme habille`e en uniformes de soleil marchant dans une jungle
\\ \hline
\multicolumn{2}{c}{\textbf{Multi30k Unsupervised}} \\ \hline
three people walking in a crowded city. & trois personnes marchant dans une rue animée. \\ \hline 
a girl with a purple shirt and sunglasses are eating. & un homme et une femme mange un plat dans un magasin local. \\ \hline
a woman sleeping in a chair with a graffiti lit street. & une femme âgée assise dans une chaise avec une canne en nuit. \\ \hline

\end{tabular}
\end{adjustwidth}
\caption{Examples of aligned generated sentences}
\label{tab:gen_sentences}
\end{table*}

% 
% \begin{table*}[h]
% k\centering
% k\small
% k\begin{tabular}{|l|l|}
% k\hline
% k\textbf{English} & \begin{tabular}[c]{@{}l@{}}what of the matter of reference to this policy is the order of community and they are not needs to\\ it unk unk , it is the unk order to meet the unk and unk it to be unk to\\ the commission needs to take unk of this situation that it has to be unk going to shape it should\\ 
% kthe european commission \&apos; s matter of political developments in europe , it is unk , european policy , so\\ 
% kthe european union unk s national order are necessary to do , but these are unk unk . unk
% k\end{tabular}     \\ \hline
% k\textbf{French}  & \begin{tabular}[c]{@{}l@{}}en tant que parlement unk , la convention devra unk davantage unk des unk de financement unk unk travers ces\\ unk de la lutte contre la unk des droits fondamentaux de ce processus est unk et unk davantage de unk\\ la unk juridique du parlement n unk est pas pour aller plus loin et unk ce que se unk .\\ 
% kla commission doit poursuivre le principe de unk , qui s unk est unk , et de unk , unk \\ 
% kpourquoi unk de la unk communautaire en ce lieu , l \&apos;union unk ne unk pas sur l unk . 
% k\end{tabular} \\ \hline
% k\end{tabular}
% k\caption{Examples of aligned generated sentences.}
% k\label{gen_sent}
% k\end{table*}

\subsection{Human Evaluation}
The subjective judgments of the generated sentences of the models trained using the Europarl and the Multi30k datasets with maximum sentence length of size 20 and 15 is reported in Table~\ref{human_eval}. As we do not have ground truth for our translation we measure parallelism between our generated sentences only based on human evaluation. We used 25 random generated sentences from each model and give them to a group of 4 bilingual people. We asked them to first rate the sentences based on a 5-point scale according to their fluency. The judges are asked to score 1 which corresponds to gibberish, 3 corresponds to understandable but ungrammatical, and 5 correspond to naturally constructed and understandable sentences~\citep{cifka}. Then, we ask ask them to measure parallelism of the generated samples assuming that the sentences are translations of each other. The scale is between 1 and 5 again with 1 corresponding to no parallelism, 3 to some parallelism and 5 to fully parallel sentences. From Table~\ref{human_eval}, we can note that on text quality human evaluation results corresponds to our other quantitative metrics. Our generated sentences show some parallelism even in the unsupervised scenario. Some example generated sentences are shown in Table~\ref{tab:gen_sentences}. As expected, sentences generated by the supervised models exhibit more parallelism compared to ones generated by unsupervised models.

% \begin{table*}[!htb]

\begin{table}[!htb]
    \small
    \centering
  \begin{tabular}{|l|c|c|c|}
  
  \multicolumn{4}{c}{\textbf{Europarl}} \\ \hline
  &  \multicolumn{2}{c|}{\textbf{Fluency}}&\multicolumn{1}{|c|}{} \\

    %\toprule
    %\multicolumn{3}{c}{Part}                   \\
    %\cmidrule{1-2}
     &  \textbf{(EN)}&\textbf{(FR)} &\textbf{Parallelism} \\ \hline
           %&\textit{F-PPL} &\textit{R-PPL} %&\textit{F-PPL} &\textit{R-PPL}   \\
    %\hline

    %\midrule
       Real   & 4.89&4.81 &4.63 \\
      Bilingual-GAN (Sup.) & 4.14  & 3.8&3.05      \\
      %Real (Unsupervised) &144.16 & - & 140.25  & - \\
      Bilingual-GAN (Unsup.) & 3.88&3.52 &2.52 \\
      \hline
      
    \multicolumn{4}{c}{\textbf{Multi30k}} \\ \hline
      Real     & 4.89 & 4.82 & 4.95\\
      Bilingual-GAN (Sup.) & 3.41 & 3.2 & 2.39 \\
      %Real (Unsupervised) &58.52 & - & 43.45  & - \\
      Bilingual-GAN (Unsup.) & 4.07 & 3.24 & 1.97\\
      \hline

    % \hline
    %\bottomrule
  \end{tabular}
  \caption{Human evaluation on the generated sentences by Bilingual-GAN using the Europarl and the Multi30k dataset.}
  
  \label{human_eval}
\end{table}

\section{Conclusion}
This work proposes a novel way of modelling NMT and NTG whereby we consider them as a joint problem from the vantage of a bilingual person. It is a step towards modeling concepts and ideas which are language agnostic using the latent representation of machine translation as the basis.  

We explore the versatility and the representation power of latent space based deep neural architectures which can align different languages and give us a principled way of generating from this shared space. Using quantitative and qualitative evaluation metrics we demonstrate that we can generate fluent sentences which exhibit parallelism in our two target languages. Future work will consist of improving the quality of the generated sentences, increasing parallelism specially without using parallel data to train the NMT and adding more languages. Other interesting extensions include using our model for conditional text generation and multi-modal tasks such as image captioning.

\bibliography{iclr2019_conference}
\bibliographystyle{acl_natbib}

\end{document}